\documentclass[10pt,twocolumn,letterpaper]{article}

\usepackage[pagenumbers]{cvpr} 

\usepackage{graphicx}
\usepackage{amsmath}
\usepackage{amssymb}
\usepackage{booktabs}
\usepackage{threeparttable}

\usepackage[pagebackref,breaklinks,colorlinks]{hyperref}

\usepackage[capitalize]{cleveref}
\crefname{section}{Sec.}{Secs.}
\Crefname{section}{Section}{Sections}
\Crefname{table}{Table}{Tables}
\crefname{table}{Tab.}{Tabs.}

\begin{document}

\title{Global Context with Discrete Diffusion in Vector Quantised Modelling \\ for Image Generation}

\author{Minghui Hu\textsuperscript{1} \and
Yujie Wang\textsuperscript{2} \and
Tat-Jen Cham \textsuperscript{1} \and
Jianfei Yang \textsuperscript{1} \and
P.N.Suganthan\textsuperscript{1} 
\and
\textsuperscript{1}{Nanyang Technological University} \and
\textsuperscript{2}{Sensetime Research} 
\and
{\tt\small \{e200008,yang0478\}@e.ntu.edu.sg} \;
{\tt\small \{astjcham,epnsugan\}@ntu.edu.sg} \and
{\tt\small wangyujie@sensetime.com}}

\maketitle

\begin{abstract}

The integration of Vector Quantised Variational AutoEncoder (VQ-VAE) with autoregressive models as generation part has yielded high-quality results on image generation. However, the autoregressive models will strictly follow the progressive scanning order during the sampling phase. This leads the existing VQ series models to hardly escape the trap of lacking global information. Denoising Diffusion Probabilistic Models (DDPM) in the continuous domain have shown a capability to capture the global context, while generating high-quality images. In the discrete state space, some works have demonstrated the potential to perform text generation and low resolution image generation. We show that with the help of a content-rich discrete visual codebook from VQ-VAE, the discrete diffusion model can also generate high fidelity images with global context, which compensates for the deficiency of the classical autoregressive model along pixel space. Meanwhile, the integration of the discrete VAE with the diffusion model resolves the drawback of conventional autoregressive models being oversized, and the diffusion model which demands excessive time in the sampling process when generating images. It is found that the quality of the generated images is heavily dependent on the discrete visual codebook. 
Extensive experiments demonstrate that the proposed Vector Quantised Discrete Diffusion Model (VQ-DDM) is able to achieve comparable performance to top-tier methods with low complexity. It also demonstrates outstanding advantages over other vectors quantised with autoregressive models in terms of image inpainting tasks without additional training.

\end{abstract}

\section{Introduction}

Vector Quantised Variational AutoEncoder (VQ-VAE) ~\cite{van2017neural} is a popular method developed to compress images into discrete representations for the generation. Typically, after the compression and discretization representation by the convolutional network, an autoregressive model is used to model and sample in the discrete latent space, including PixelCNN family~\cite{oord2016conditional,van2016pixel,chen2018pixelsnail}, transformers family~\cite{ramesh2021zero,chen2020generative}, etc. 
However, in addition to the disadvantage of the huge number of model parameters, these autoregressive models can only make predictions based on the observed pixels (left upper part of the target pixel) due to the inductive bias caused by the strict adherence to the progressive scan order~\cite{khan2021transformers,bengio2015scheduled}. If the conditional information is located at the end of the autoregressive sequence, it is difficult for the model to obtain relevant information. 
\begin{figure}
    \centering
    \includegraphics[scale=0.55]{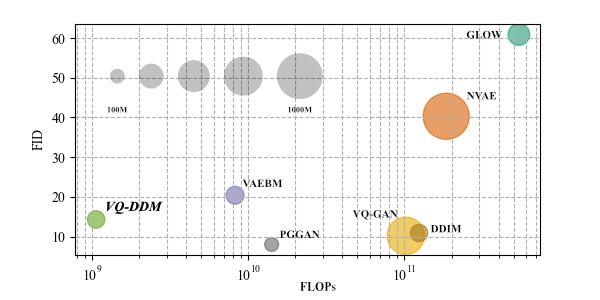}
    \caption{FID v.s. Operations and Parameters. The size of the blobs is proportional to the number of network parameters, the X-axis indicates FLOPs on a log scale and the Y-axis is the FID score.}
    \label{fig1}
\end{figure}

A recent alternative generative model is the Denoising Diffusion Model, which can effectively mitigate the lack of global information ~\cite{sohl2015deep,ho2020denoising}, also achieving comparable or state-of-the-art performance in text~\cite{hoogeboom2021argmax,austin2021structured}, image~\cite{dhariwal2021diffusion} and speech generation~\cite{kong2020diffwave} tasks. Diffusion models are parameterized Markov chains trained to translate simple distributions to more sophisticated target data distributions in a finite set of steps. Typically the Markov chain begins with an isotropic Gaussian distribution in continuous state space, with the transitions of the chain for reversing a diffusion process that gradually adds Gaussian noise to source images. In the inverse process, as the current step is based on the global information of the previous step in the chain, this endows the diffusion model with the ability to capture the global information.

However, the diffusion model has a non-negligible disadvantage in that the time and computational effort involved in generating the images are enormous. The main reason is that the reverse process typically contains thousands of steps. Although we do not need to iterate through all the steps when training, all these steps are still required when generating a sample, which is much slower compared to GANs and even autoregressive models. 
 
Some recent works~\cite{song2020denoising,nichol2021improved} have attempted addressing these issues by decreasing the sampling steps, but the computation cost is still high as each step of the reverse process generates a full-resolution image.

In this work, we propose the \textbf{V}ector \textbf{Q}uantized \textbf{D}iscrete \textbf{D}iffusion \textbf{M}odel (VQ-DDM), a versatile framework for image generation consisting of a discrete variational autoencoder and a discrete diffusion model. VQ-DDM consists of two stages: (1) learning an abundant and efficient discrete representation of images, (2) fitting the prior distribution of such latent visual codes via discrete diffusion model.

VQ-DDM substantially reduces the computational resources and required time to generate high-resolution images by using a discrete scheme. Then the common problem of the lack of global content and overly large number of parameters of the autoregressive model is solved by fitting a latent variable prior using the discrete diffusion model. Finally, since a bias of codebook will limit generation quality, while model size is also dependent on the number of categories, we propose a re-build and fine-tune(ReFiT) strategy to construct a codebook with higher utilization, which will also reduce the number of parameters in our model.

In summary, our key contributions include the following:
\begin{itemize}

    \item VQ-DDM fits the prior over discrete latent codes with a discrete diffusion model. The use of diffusion model allows the generative models consider the global information instead of only focusing on partially seen context to avoid sequential bias.

    \item We propose a ReFiT approach to improve the utilisation of latent representations in the visual codebook, which can increase the code usage of VQ-GAN from $31.85\%$ to $97.07\%$, while the FID between reconstruction image and original training image is reduced from $10.18$ to $5.64$ on CelebA-HQ $256\times256$.

    \item VQ-DDM is highly efficient for the both number of parameters and generation speed. As shown in Figure~\ref{fig1}, using only 120M parameters, it outperforms VQ-VAE-2 with around 10B parameters and is comparable with VQ-GAN with 1B parameters in image generation tasks in terms of image quality. It is also 10 $ \sim $ 100 times faster than other diffusion models for image generation~\cite{song2020denoising,ho2020denoising}.

\end{itemize}

\begin{figure*}[h]
    \centering
    \includegraphics[scale=0.15]{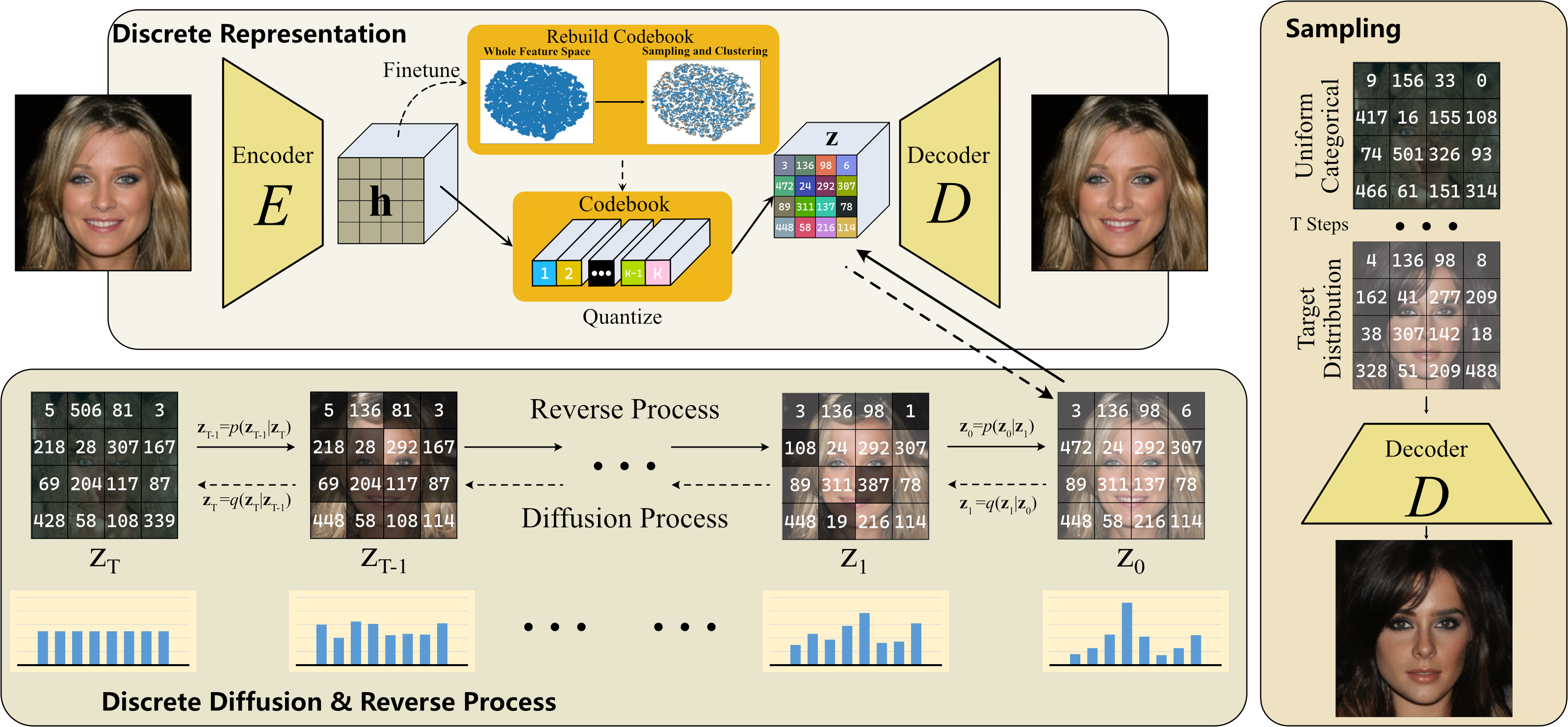}
    \caption{The proposed VQ-DDM pipeline contains 2 stages: (1) Compress the image into discrete variables via discrete VAE. (2) Fit a prior distribution over discrete coding by a diffusion model.
    Black squares in the diffusion diagram illustrate states when the underlying distributions are uninformative, but which become progressively more specific during the reverse process.
    The bar chart at the bottom of the image represents the probability of a particular discrete variable being sampled.
    }
    \label{pipeline}
\end{figure*}

\section{Preliminaries}
\subsection{Diffusion Models in continuous state space}

Given data $\mathbf{x}_0$ from a data distribution $q(\mathbf{x}_0)$, the diffusion model consists of two processes: the \textit{diffusion process} and the \textit{reverse process}~\cite{sohl2015deep,ho2020denoising}. 

The \textit{diffusion process} progressively destroys the data $\mathbf{x}_0$ into $\mathbf{x}_T$ over $T$ steps, via a fixed Markov chain that gradually introduces Gaussian noise to the data according to a variance schedule $\beta_{1:T} \in (0,1]^T$ as follows:
\begin{equation}
    q(\mathbf{x}_{1:T}|\mathbf{x}_0) = \prod_{t=1}^T q(\mathbf{x}_t|\mathbf{x}_{t-1}) ,
\end{equation}
\begin{equation}
    q(\mathbf{x}_t | \mathbf{x}_{t-1}) = \mathcal{N}(\mathbf{x}_t;\sqrt{1-\beta_t}\mathbf{x}_{t-1},\beta_t \mathbf{I}) .
\end{equation}

With an adequate number of steps $T$ and a suitable variance schedule $\beta$,
$p(\mathbf{x}_T)$ becomes an isotropic Gaussian distribution.

The \textit{reverse process} is defined as a Markov chain parameterized by $\theta$, which is used to restore the data from the noise:
\begin{equation}
    p_{\theta}(\mathbf{x}_{0:T}) = p(\mathbf{x}_T) \prod_{t=1}^T p_{\theta} (\mathbf{x}_{t-1}|\mathbf{x}_t),
\end{equation}
\begin{equation}
    p_{\theta}(\mathbf{x}_{t-1}|\mathbf{x}_{t}) = \mathcal{N} (\mathbf{x}_{t-1};\mu_{\theta}(\mathbf{x}_t,t),\Sigma_{\theta}(\mathbf{x}_t,t)).
\end{equation}

The objective of training is to find the best $\theta$ to fit the data distribution $q(\mathbf{x}_0)$ by optimizing the variational lower bound (VLB)~\cite{kingma2013auto}
\begin{equation}
\begin{split}
        \mathbb{E}_{q(\mathbf{x}_0)}& [\log p_{\theta}(\mathbf{x}_0)]\\ = &\mathbb{E}_{q(\mathbf{x}_0)}\log\mathbb{E}_{q(\mathbf{x}_{1:T}|\mathbf{x}_0)} \left[ \frac{p_{\theta}(\mathbf{x}_{0:T})}{q(\mathbf{x}_{1:T}|\mathbf{x}_0)} \right] \\ \geq &\mathbb{E}_{q(\mathbf{x}_{0:T})} \left[ \log \frac{p_{\theta}(\mathbf{x}_{0:T})}{q(\mathbf{x}_{1:T}|\mathbf{x}_0)} \right]  =: L_\mathrm{vlb}.
\end{split}
\label{vlb}
\end{equation}

Ho \etal \cite{ho2020denoising} revealed that the variational lower bound in Eq.~\ref{vlb} can be calculated with closed form expressions instead of Monte Carlo estimates as the \textit{diffusion process} posteriors and marginals are Gaussian, which allows sampling $\mathbf{x}_t$ at an arbitrary step $t$ with $\alpha_t = 1-\beta_t$,  $\bar{\alpha}_t=\prod_{s=0}^t \alpha_s$ and  $\tilde{\beta_t}=\frac{1-\bar{\alpha}_{t-1}}{1-\bar{\alpha}_t}$:
\begin{equation}
    q(\mathbf{x}_t|\mathbf{x}_0) = \mathcal{N}(\mathbf{x}_t | \sqrt{\bar{\alpha}_t} \mathbf{x}_0, (1-\bar{\alpha}_t)\mathbf{I} ),
\end{equation}

\begin{equation}
\begin{split}
    L_\mathrm{vlb} =  \mathbb{E}_{q(\mathbf{x}_0)} &[ D_{\mathrm{KL}}(q(\mathbf{x}_T|\mathbf{x}_0) || p(\mathbf{x}_T)) - \log p_{\theta} ( \mathbf{x}_0 | \mathbf{x}_1 ) \\
    &+ \sum_{t=2}^T D_{\mathrm{KL}}(q(\mathbf{x}_{t-1}|\mathbf{x}_t,\mathbf{x}_0) || p_{\theta}(\mathbf{x}_{t-1}|\mathbf{x}_t)) ].
\end{split}
\label{kl}
\end{equation}

Thus the reverse process can be parameterized by neural networks $\epsilon_{\theta}$ and $\upsilon_{\theta}$, which can be defined as:
\begin{equation}
    \mu_{\theta}(\mathbf{x}_t,t) = \frac{1}{\sqrt{\alpha_t}} \left(\mathbf{x}_t - \frac{\beta_t}{\sqrt{1-\bar{\alpha}_t}} \epsilon_{\theta} (\mathbf{x}_t,t) \right),
\end{equation}
\begin{equation}
\begin{split}
    \Sigma_{\theta}(\mathbf{x}_t,t) = \exp(\upsilon_{\theta}&(\mathbf{x}_t,t)\log\beta_t \\
    &+ (1-\upsilon_{\theta}(\mathbf{x}_t,t))\log\tilde{\beta_t}).
\end{split}
\end{equation}

Using a modified variant of the VLB loss as a simple loss function will offer better results in the case of fixed $\Sigma_{\theta}$~\cite{ho2020denoising}:
\begin{equation}
    L_{\mathrm{simple}} = \mathbb{E}_{t,\mathbf{x}_0,\epsilon} \left[ ||\epsilon - \epsilon_{\theta}(\mathbf{x}_t,t)||^2 \right],
\end{equation}
which is a reweighted version resembling denoising score matching over multiple noise scales indexed by $t$~\cite{song2019generative}.

Nichol \etal \cite{nichol2021improved} used an additional $L_{\mathrm{vlb}}$ to the simple loss for guiding a learned $\Sigma_{\theta}(\mathbf{x}_t,t)$, while keeping the $\mu_{\theta}(\mathbf{x}_t,t)$ still the dominant component of the total loss:
\begin{equation}
    L_{\mathrm{hybrid}} = L_{\mathrm{simple}} + \lambda L_{\mathrm{vlb}}.
\end{equation}

\subsection{Discrete Representation of Images}
van den Oord \etal \cite{van2017neural} presented a discrete variational autoencoder with a categorical distribution as the latent prior, which is able to map the images into a sequence of discrete latent variables by an encoder and reconstruct the image according to those variables with a decoder.
Formally, given a codebook $\mathbb{Z}\in\mathbb{R}^{K\times d}$, where $K$ represents the capacity of latent variables in the codebook and $d$ is the dimension of each latent variable, after compressing the high dimension input data $\textbf{x}\in \mathbb{R}^{c\times H\times W}$ into latent vectors $\textbf{h}\in \mathbb{R}^{h\times w\times d}$ by an encoder $E$, $\textbf{z}$ is the quantised $\textbf{h}$, which substitutes the vectors $h_{i,j}\in\textbf{h}$ by the nearest neighbor $z_k \in \mathbb{Z}$. The decoder $D$ is trained to reconstruct the data from the quantised encoding $\textbf{z}_q$:
\begin{equation}
    \textbf{z} = \mathrm{Quantize}(\textbf{h}) :=  \mathrm{arg\ min}_k ||h_{i,j}-z_k|| ,
\end{equation}
\begin{equation}
    \hat{\textbf{x}} = D(\textbf{z}) = D(\mathrm{Quantize}(E(\textbf{x}))).
\end{equation}

As $\mathrm{Quantize}(\cdot)$ has a non-differentiable operation $\mathrm{arg\ min}$, the straight-through gradient estimator is used for back-propagating the reconstruction error from decoder to encoder. The whole model can be trained in an end-to-end manner by minimizing the following function:
\begin{equation}
    L = ||\textbf{x}-\hat{\textbf{x}}||^2 + ||sg[E(\textbf{x})] - \textbf{z}|| + \beta || sg[\textbf{z}] - E(\textbf{x}) || ,
\label{vqeq}
\end{equation}
where $sg[\cdot]$ denotes stop gradient and broadly the three terms are reconstruction loss, codebook loss and commitment loss, respectively. 
VQ-GAN~\cite{esser2021taming} extends VQ-VAE~\cite{van2017neural} in multiple ways. It substitutes the L1 or L2 loss of the original VQ-VAE with a perceptual loss~\cite{zhang2018unreasonable}, and adds an additional discriminator to distinguish between real and generated patches~\cite{CycleGAN2017}.

The codebook update of the discrete variational autoencoder is intrinsically a dictionary learning process. Its objective uses L2 loss to narrow the gap between the codes $\mathbb{Z}_t \in \mathbb{R}^{K_t\times d}$ and the encoder output $\textbf{h}\in\mathbb{R}^{ h\times w \times d}$ \cite{van2017neural}. In other words, the codebook training is like $k$-means clustering, where cluster centers are the discrete latent codes. However, since the volume of the codebook space is dimensionless and $\textbf{h}$ is updated each iteration, the discrete codes $\mathbb{Z}$ typically do not follow the encoder training quickly enough. Only a few codes get updated during training, with most unused after initialization.

\section{Methods}
Our goal is to leverage the powerful generative capability of the diffusion model to perform high fidelity image generation tasks with a low number of parameters. 

Our proposed method, VQ-DDM, is capable of generating high fidelity images with a relatively small number of parameters and FLOPs, as summarized in Figure ~\ref{pipeline}.
Our solution starts by compressing the image into discrete variables via the discrete VAE and then constructs a powerful model to fit the joint distribution over the discrete codes by a diffusion model. During diffusion training, the darker coloured parts in Figure ~\ref{pipeline} represent noise introduced by uniform resampling. When the last moment is reached, the latent codes have been completely corrupted into noise. 
In the sampling phase, the latent codes are drawn from an uniform categorical distribution at first, and then resampled by performing reverse process $T$ steps to get the target latent codes. Eventually, target latent codes are pushed into the decoder to generate the image.

\subsection{Discrete Diffusion Model} \label{ddm}

Assume the discretization is done with $K$ categories, i.e.\
$z_t \in \{1,\dots,K\}$, with the one-hot vector representation given by $\textbf{z}_t \in \{0,1\}^K$. The corresponding probability distribution is expressed by $\textbf{z}_t^{\mathrm{logits}}$ in logits. We formulate the discrete diffusion process as
\begin{equation}
    q(\textbf{z}_t|\textbf{z}_{t-1}) = \mathrm{Cat} (\textbf{z}_t ; \textbf{z}_{t-1}^{\mathrm{logits}} \mathbf{Q}_t  ),
\end{equation}
where $\mathrm{Cat}(\textbf{x}|\textbf{p})$ is the categorical distribution parameterized by $\textbf{p}$, while $\mathbf{Q}_t$ is the process transition matrix. In our method, $\mathbf{Q}_t = (1-\beta_t)\textbf{I} + \beta_t / K $, which means $\textbf{z}_t$ has $1-\beta_t$ probability to keep the state from last timestep and $\beta_t$ chance to resample from a uniform categorical distribution. Formally, it can be written as 
\begin{equation}
    q(\textbf{z}_t|\textbf{z}_{t-1}) = \mathrm{Cat} (\textbf{z}_t ; (1-\beta_t)\textbf{z}_{t-1}^{\mathrm{logits}} + \beta_t / K).
\label{ddp}
\end{equation}

It is straightforward to get $\textbf{z}_t$ from $\textbf{z}_0$ under the schedule $\beta_t$ with $\alpha_t = 1-\beta_t$,  $\bar{\alpha}_t=\prod_{s=0}^t \alpha_s$:
\begin{equation}
    q(\textbf{z}_t|\textbf{z}_{0}) = \mathrm{Cat}(\textbf{z}_t ; \bar{\alpha}_t \textbf{z}_0 + (1-\bar{\alpha}_t)/K)
\label{ddp0}
\end{equation}
\begin{equation}
    or \quad  q(\textbf{z}_t|\textbf{z}_{0}) = \mathrm{Cat}(\textbf{z}_t ; \textbf{z}_0 \bar{\mathbf{Q}}_t) ; \  \bar{\mathbf{Q}}_t = \prod_{s=0}^t \mathbf{Q}_s.
\end{equation}

We use the same cosine noise schedule as \cite{nichol2021improved,hoogeboom2021argmax} because our discrete model is also established on the latent codes with a small $16\times16$ resolution. Mathematically, it can be expressed in the case of $\bar{\alpha}$ by 
\begin{equation}
    \bar{\alpha} = \frac{f(t)}{f(0)}, \quad f(t) =\mathrm{cos}\left(\frac{t/T+s}{1+s} \times \frac{\pi}{2}\right)^2 .
\label{noises}
\end{equation}

By applying Bayes' rule, we can compute the posterior $q(\textbf{z}_{t-1}|\textbf{z}_{t},\textbf{z}_{0})$ as:

\begin{equation}
    \begin{split}
    q(\textbf{z}_{t-1} |  \textbf{z}_{t},\textbf{z}_{0})& = \mathrm{Cat} \left(\textbf{z}_t ; \frac{\textbf{z}_t^{\mathrm{logits}} \mathbf{Q}_t^{\top} \odot \textbf{z}_0 \bar{\mathbf{Q}}_{t-1} }{\textbf{z}_0 \bar{\mathbf{Q}}_{t} {\textbf{z}_t^{\mathrm{logits}}}^{\top}} \right) \\ 
    =  \mathrm{Cat} &(\textbf{z}_t ;  \ \boldsymbol{\theta}(\textbf{z}_t,\textbf{z}_0) / \sum_{k=1}^K \theta_k (z_{t,k},z_{0,k}) ), \\
    \end{split}
    \label{qpost}
\end{equation}
\begin{equation}
\begin{split}
            \boldsymbol{\theta}(\textbf{z}_t,\textbf{z}_0)  = [\alpha_t \textbf{z}_t^{\mathrm{logits}} + & (1-\alpha_t)/ K]  \\ &\odot [\bar{\alpha}_{t-1} \textbf{z}_0 + (1-\bar{\alpha}_{t-1}) / K].
\end{split}
\end{equation}

It is worth noting that $ \boldsymbol{\theta}(\textbf{z}_t,\textbf{z}_0) / \sum_{k=1}^K \theta_k (z_{t,k},z_{0,k})$ is the normalized version of $\boldsymbol{\theta}(\textbf{z}_t,\textbf{z}_0)$, and we use $\mathrm{N}[\boldsymbol{\theta}(\textbf{z}_t,\textbf{z}_0)]$ to denote $ \boldsymbol{\theta}(\textbf{z}_t,\textbf{z}_0) / \sum_{k=1}^K \theta_k (z_{t,k},z_{0,k})$ below.

Hoogeboom \etal \cite{hoogeboom2021argmax} predicted $\hat{\textbf{z}}_0$ from $\textbf{z}_t$ with a neural network $\mu(\textbf{z}_t,t)$, instead of directly predicting $p_{\theta}(\textbf{z}_{t-1}|\textbf{z}_{t})$. Thus the reverse process can be parameterized by the probability vector from  $q(\textbf{z}_{t-1} |  \textbf{z}_{t},\hat{\textbf{z}}_{0})$. Generally, the reverse process $p_{\theta}(\textbf{z}_{t-1}|\textbf{z}_{t})$ can be expressed by
\begin{equation}
\begin{split}
        p_{\theta}(\textbf{z}_0|\textbf{z}_1) & = \mathrm{Cat} (\textbf{z}_0 |\hat{\textbf{z}}_0), \\ 
        p_{\theta}(\textbf{z}_{t-1}|\textbf{z}_{t})& = \mathrm{Cat}  (\textbf{z}_t | \  \mathrm{N}[\boldsymbol{\theta}(\textbf{z}_t,\hat{\textbf{z}}_0)]) .
\end{split}
\end{equation}

Inspired by~\cite{jang2016categorical,maddison2016concrete}, we use a neural network $\mu(\mathbf{Z}_t,t)$ to learn and predict the a noise $n_t$ and obtain the logits of $\hat{\mathbf{z}}_0$ from
\begin{equation}
    \hat{\mathbf{z}}_0 = \mu(\mathbf{Z}_t,t) + \mathbf{Z}_t.
    \label{pnois}
\end{equation}

It is worth noting that the neural network $\mu(\cdot)$ is based on the  $\mathbf{Z}_t \in \mathbb{N}^{h\times w}$, where all the discrete representation $\mathbf{z}_t$ of the image are combined. The final noise prior $\mathbf{Z}_T$ is uninformative, and it is possible to separably sample from each axis during inference. However, the reverse process is jointly informed and evolves towards a highly coupled $\mathbf{Z}_0$. We do not define a specific joint prior for $\mathbf{z}_t$, but encode the joint relationship into the learned reverse process. This is implicitly done in the continuous domain diffusion. As $\mathbf{z}_{t-1}$ is based on the whole previous representation $\mathbf{z}_t$, the reverse process can sample the whole discrete code map directly while capturing the global information.

The loss function used is the VLB from Eq.~\ref{kl}, where the summed KL 
divergence for $T>2$ is given by

\begin{equation}
\begin{split}
    \mathrm{KL}( q(\textbf{z}_{t-1} |  \textbf{z}_{t},\textbf{z}_{0}) || p_{\theta}(\textbf{z}_{t-1}|\textbf{z}_{t})) &= \\
    \sum_k  \mathrm{N}[\boldsymbol{\theta}(\textbf{z}_t,\textbf{z}_0)]  &\times
    \log \frac{\mathrm{N}[\boldsymbol{\theta}(\textbf{z}_t,\textbf{z}_0)] }{\mathrm{N}[\boldsymbol{\theta}(\textbf{z}_t,\hat{\textbf{z}}_0)] }.
\end{split}
\end{equation}

\subsection{Re-build and Fine-tune Strategy}
Our discrete diffusion model is based on the latent representation of the discrete VAE codebook $\mathbb{Z}$. However, the codebooks with rich content are normally large, with some even reaching $K=16384$. This makes it highly unwieldy for our discrete diffusion model, as the transition matrices of discrete diffusion models have a quadratic level of growth to the number of classes $K$, \eg $O(K^2T)$~\cite{austin2021structured}.

To reduce the categories used for our diffusion model, we proposed a Re-build and Fine-tune (ReFit) strategy to decrease the size $K$ of codebook $\mathbb{Z}$ and boost the reconstruction performance based on a well-trained discrete VAEs trained by the straight-through method.

From Eq.~\ref{vqeq}, we can find the second term and the third term are related to the codebook, but only the second term is involved in the update of the codebook. $||sg[E(\textbf{x})] - \textbf{z}||$ reveals that only a few selected codes, the same number as the features from $E(\textbf{x})$, are engaged in the update per iteration. Most of the codes are not updated or used after initialization, and the update of the codebook can lapse into a local optimum. 

We introduce a re-build and fine-tune strategy to avoid the waste of codebook capacity. With the trained encoder, we reconstruct the codebook so that all codes in the codebook have the opportunity to be selected. This will greatly increase the usage of the codebook. Suppose we desire to obtain a discrete VAE having a codebook with $\mathbb{Z}_t$ based on a trained discrete VAE with an encoder $E_s$ and a decoder $D_s$. We first encode each image $\textbf{x}\in \mathbb{R}^{c\times H\times W}$ to latent features $\textbf{h}$, or loosely speaking, each image gives us $h\times w$ features with $d$ dimension. Next we sample $P$ features uniformly
from the entire set of features found in training images,
where $P$ is the sampling number and far larger than the desired codebook capacity $K_t$. This ensures that the re-build codebook is composed of valid latent codes. Since the process of codebook training is basically the process of finding cluster centres, we directly employ k-means with AFK-MC$^2$~\cite{bachem2016fast} on the sampled $P$ features and utilize the centres to re-build the codebook $\mathbb{Z}_t$. We then replace the original codebook with the re-build $\mathbb{Z}_t$ and fine-tune it on top of the well-trained discrete VAE.

\section{Experiments and Analysis}

\subsection{Datasets and Implementation Details} \label{desc}

We show the effectiveness of the proposed VQ-DDM on \textit{CelebA-HQ}~\cite{karras2017progressive} and \textit{LSUN-Church}~\cite{yu2015lsun} datasets and verify the proposed Re-build and Fine-tune strategy on \textit{CelebA-HQ} and \textit{ImageNet} datasets. The details of the dataset are given in the Appendix.

The discrete VAE follows the same training strategy as VQ-GAN\cite{esser2021taming}. All training images are processed to $256\times256$, and the compress ratio is set to $16$, which means the latent vector $\textbf{z} \in \mathbb{R}^{1\times16\times16}$. 
When conducting Rebuild and Fine-tune, the sampling number $P$ is set to $20k$ for \textit{LSUN} and \textit{CelebA}. For the more content-rich case, we tried a larger P value $50k$ for \textit{ImageNet}. In practical experiments, we sample $P$ images with replacement uniformly from the whole training data and obtained corresponding latent features. For each feature map, we make another uniform sampling over the feature map size $16\times16$ to get the desired features. In the fine-tuning phase, we freeze the encoder and set the learning rate of the decoder to $1e$-$6$ and the learning rate of the discriminator to $2e$-$6$ with 8 instances per batch.

With regard to the diffusion model, the network for estimating $n_t$ has the same structure as~\cite{ho2020denoising}, which is a U-Net~\cite{ronneberger2015u} with self-attention~\cite{vaswani2017attention}. The detailed settings of hyperparameters are provided in the Appendix. We set timestep $T=4000$ in our experiments and the noise schedule is the same as~\cite{nichol2021improved}

\subsection{Codebook Quality} \label{cbq}
A large codebook dramatically increases the cost of DDM.
To reduce the cost to an acceptable scale, we proposed a resample and fine-tune strategy to compress the size of the codebook, while maintaining quality.
To demonstrate the effectiveness of the proposed strategy, we compare the codebook usage and FID of reconstructed images of our method to VQ-GAN\cite{esser2021taming}, VQ-VAE-2\cite{razavi2019generating} and DALL-E\cite{ramesh2021zero}. 

In this experiment, we compressed the images from $3\times256\times256$ to $1\times16\times16$ with two different codebook capacities $K=\{512,1024\}$. 
We also proposed an indicator to measure the usage rate of the codebook, which is the number of discrete features that have appeared in the test set or training set divided by the codebook capacity.

\begin{figure*}
    \centering
    \includegraphics[scale=0.20]{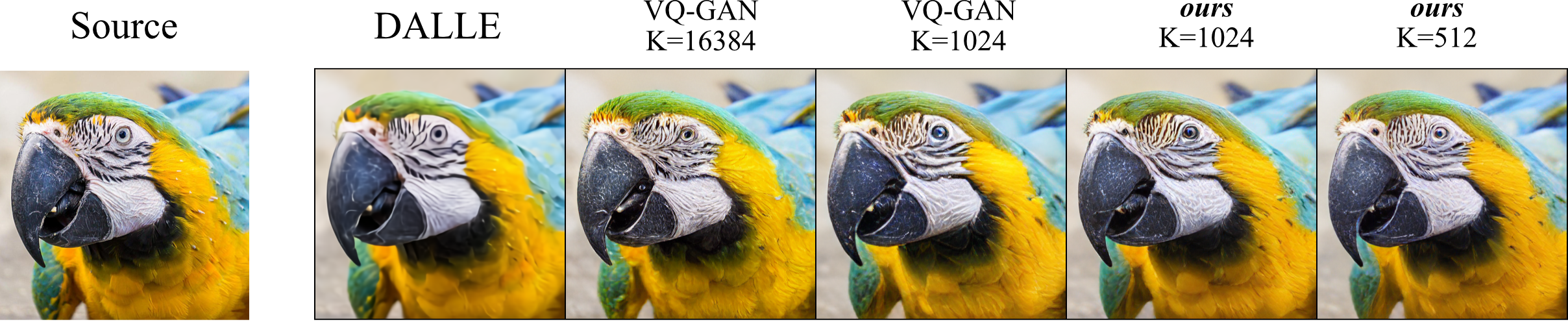}
    \caption{Reconstruction images $384\times384$ from ImageNet based VQ-GAN and ReFiT}
    \label{Recon-inr}
\end{figure*}

\begin{figure}[t!]
\centering
\resizebox{0.50\textwidth}{!}{
\begin{subfigure}{0.125\textwidth}
    \centering
    \includegraphics[scale=0.20]{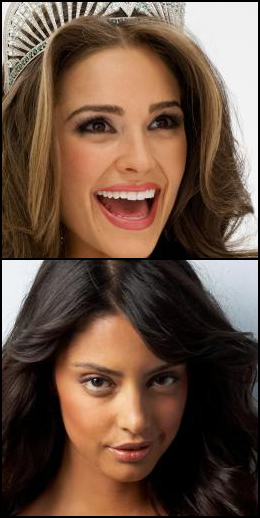} 
    \caption{Source}
  \end{subfigure}
 \begin{subfigure}{0.125\textwidth}
    \centering
    \includegraphics[scale=0.20]{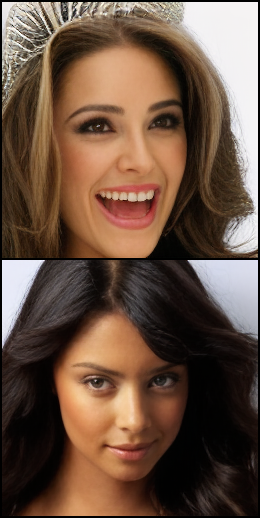} 
    \caption{VQ-GAN}
  \end{subfigure}
  
 \begin{subfigure}{0.125\textwidth}
    \centering
    \includegraphics[scale=0.20]{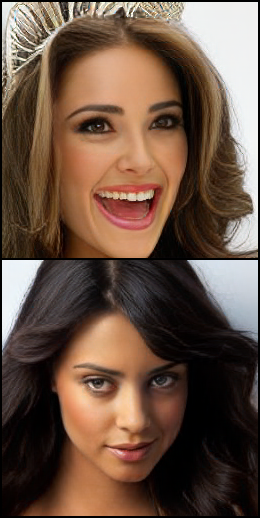} 
    \caption{ReFiT K=1024}
  \end{subfigure}
  
 \begin{subfigure}{0.125\textwidth}
    \centering
    \includegraphics[scale=0.20]{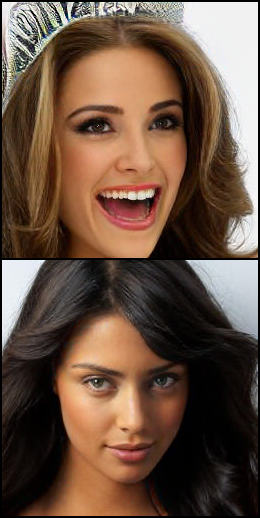} 
    \caption{ReFiT K=512}
  \end{subfigure}
}

  \caption{Reconstruction images of CelebA HQ $256\times256$ from VQ-GAN and ReFiT.}
  \label{Recon-celeba}
\end{figure}

The quantitative comparison results are shown in Table~\ref{codebook_com} while the reconstruct images are demonstrated in Figs.~\ref{Recon-inr} \& \ref{Recon-celeba}.
Reducing the codebook capacity from 1024 to 512 only brings $\sim 0.1$ decline in CelebA and $\sim 1$ in ImageNet. 
As seen in Figure ~\ref{Recon-celeba}, the reconstructed images (c,d) after ReFiT strategy are richer in colour and more realistic in expression than the reconstructions from VQ-GAN (b).
The codebook usage of our method has improved significantly compared to other methods, nearly 3x high than the second best. 
Our method also achieves the equivalent reconstruction quality at the same compression rate and with 32$\times$ lower capacity $K$ of codebook $\mathbb{Z}$.

For VQ-GAN with capacity $16384$, although it only has $976$ effective codes, which is smaller than $1024$ in our ReFiT method when $P=20k$, it achieves a lower FID in reconstructed images vs validation images. One possible reason is that the value of $P$ is not large enough to cover some infrequent combinations of features during the re-build phase. As the results in Table~\ref{codebook_com}, after we increase the sampling number $P$ from $20k$ to $100k$, we observe that increasing the value of $P$  achieved higher performance.

\begin{table}
\centering
\resizebox{0.46\textwidth}{!}
{%
\begin{threeparttable}
\begin{tabular}{ccccccc}
\toprule
Model &Latent Size & Capacity & \multicolumn{2}{c}{Usage of $\mathbb{Z}$} & \multicolumn{2}{c}{FID  $\downarrow$} \\ 
                       &                              &                           & CelebA      & ImageNet    & CelebA    & ImageNet    \\\midrule
VQ-VAE-2               & Cascade                      & 512                       & $\sim$65\%           &-            & -         & $\sim$10    \\
DALL-E                 & 32x32                        & 8192                      & -           & -           & -         & 32.01       \\
VQ-GAN                 & 16x16                        & 16384                     & -           & 5.96\%      & -         & 4.98        \\
VQ-GAN                 & 16x16                        & 1024                      & 31.85\%     & 33.67\%     & 10.18     & 7.94        \\
\textit{\textbf{ours}} ($P=100k$)& 16x16                        & 1024                       & -     & 100\%       & -      &4.98           \\
\textit{\textbf{ours}} ($P=20k$)& 16x16                        & 1024                      & 97.07\%     & 100\%       & 5.59          & 5.99        \\
\textit{\textbf{ours}} ($P=20k$) & 16x16                        & 512                       & 93.06\%     & 100\%       & 5.64      &6.95            \\ \bottomrule
\end{tabular}%
      \begin{tablenotes}
    \item[1] All methods are trained straight-through, except DALL-E
    with Gumbel-Softmax~\cite{ramesh2021zero}.
    \item[2] CelebA-HQ at $256$$\times$$256$. Reported FID is between 30$k$ reconstructed data vs training data.
    \item[3] Reported FID is between 50$k$ reconstructed data vs validation data
  \end{tablenotes}
\end{threeparttable}
}
    \caption{FID between reconstructed images and original images on CelebA-HQ and ImageNet }
    \label{codebook_com}
\end{table}

\subsection{Generation Quality} \label{genq}
We evaluate the performance of VQ-DDM for the unconditional image generation on \textit{CelebA-HQ} $256\times256$. Specifically, we evaluated the performance of our approach in terms of FID and compared it with various likelihood-based based methods including GLOW~\cite{kingma2018glow},  NVAE~\cite{vahdat2020nvae}, VAEBM~\cite{xiao2020vaebm}, DC-VAE~\cite{parmar2021dual}, VQ-GAN~\cite{esser2021taming} and likelihood-free method, e.g., PGGAN~\cite{karras2017progressive}. We also conducted an experiment on \textit{LSUN-Church}.

In \textit{CelebA-HQ} experiments, the discrete diffusion model was trained with $K=512$ and $K=1024$ codebooks respectively. We also report the different FID from $T=2$ to $T=4000$ with corresponding time consumption in Figure ~\ref{cost}. Regarding the generation speed, it took about 1000 hours to generate $50k$ $256\times256$ images using DDPM with 1000 steps on a NVIDIA 2080Ti GPU, 100 hours for DDIM with 100 steps~\cite{song2020denoising}, and around 10 hours for our VQ-DDM with 1000 steps.

\begin{figure}
    \centering
    \includegraphics[scale=.5]{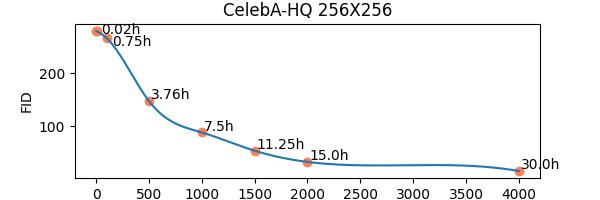}
    \caption{Steps and corresponding FID during the sampling. The text annotations are hours to sample 50k latent feature maps on 1 NVIDIA 2080Ti GPU}
    \end{figure}

\begin{figure}
    \centering
    \includegraphics[scale=.4]{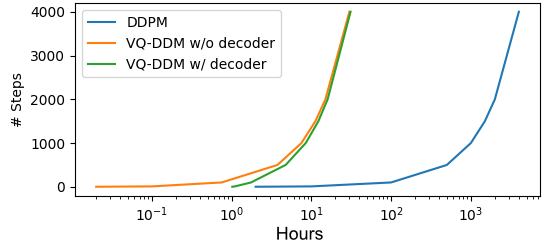}
    \caption{Hours to sampling 50k latent codes by VQ-DDM and generating 50k images with VQ-DDM and DDPM}
    \label{cost}
\end{figure}

\begin{figure*}[t!]
    \centering
    \subcaptionbox{Samples $(256\times256)$ from a VQ-DDM model trained on CelebA HQ. FID=$13.2$ \label{celebs}}{
    \includegraphics[scale=0.27]{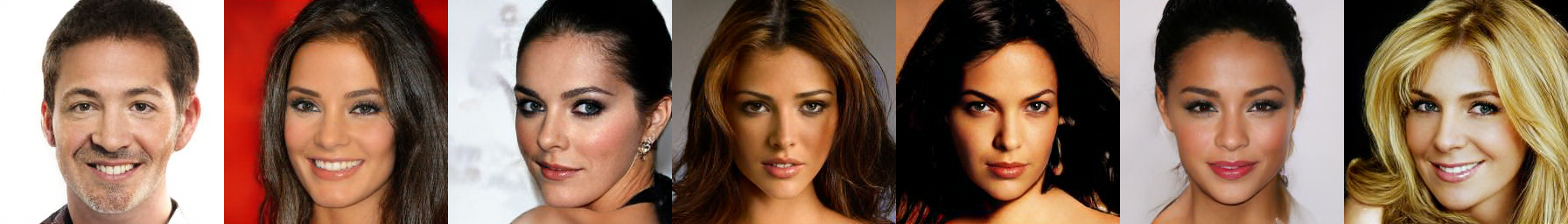}
    }

    \subcaptionbox{Samples $(256\times256)$ from a VQ-DDM model trained on LSUN-Church. FID=$16.9$ \label{lsuns}}{
    \includegraphics[scale=0.27]{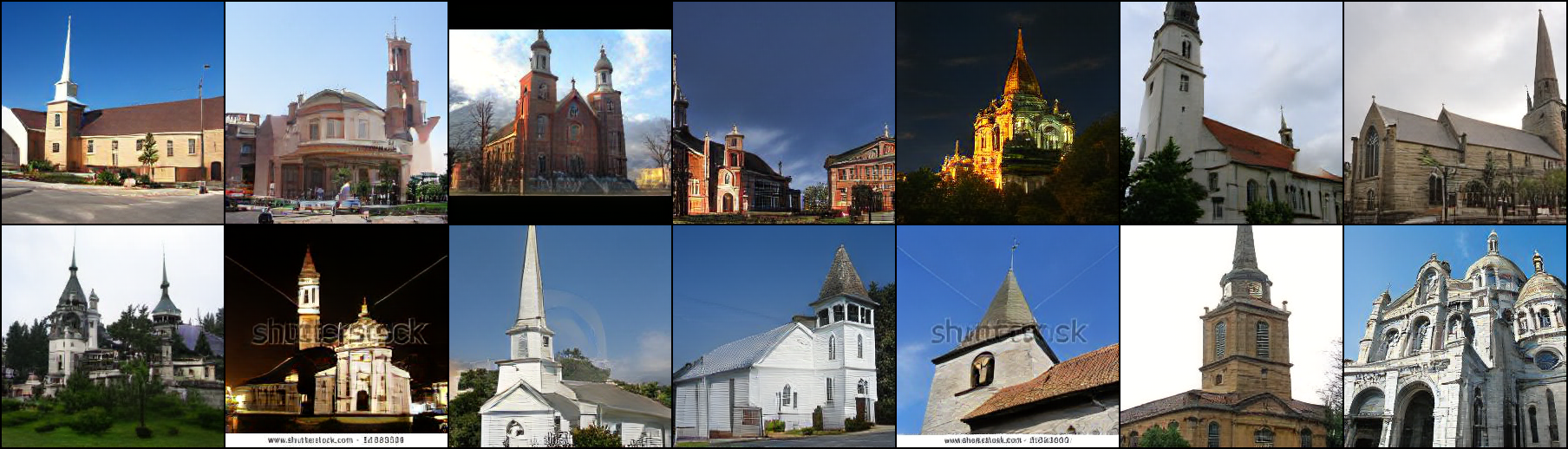}
    }
    \caption{Samples from VQ-DDM models.}
\end{figure*}

Table~\ref{celeba} shows the main results on VQ-DDM along with other established models. Although VQ-DDM is also a likelihood-based method, the training phase relies on the negative log-likehood (NLL) of discrete hidden variables, so we do not compare the NLL between our method and the other methods. The training NLL is around $1.258$ and test NLL is $1.286$ while the FID is $13.2$. Fig.~\ref{celebs} shows the generated samples from VQ-DDM trained on the \textit{CelebA-HQ}. 

For \textit{LSUN-Church}, the codebook capacity $K$ is set to $1024$, while the other parameters are set exactly the same. The training NLL is $1.803$ and the test NLL is $1.756$ while the FID between the generated images and the training set is $16.9$. Some samples are shown in Fig.~\ref{lsuns}.

After utilizing ReFiT, the generation quality of the model is significantly improved, which implies a decent codebook can have a significant impact on the subsequent generative phase. Within a certain range, the larger the codebook capacity leads to a better performance. However, excessive number of codebook entries will cause the model collapse~\cite{hoogeboom2021argmax}.

\subsection{Image Inpainting} \label{gbq}

Autoregressive models have recently demonstrated superior performance in the image inpainting tasks~\cite{chen2020generative, esser2021taming}. 
However, one limitation of this approach is that if the important context is found at the end of the autoregressive series, the models will not be able to correctly complete the images. As mentioned in Sec.~\ref{ddm}, the diffusion model will directly sample the full latent code map, with sampling steps based on the \emph{full} discrete map of the previous step. Hence it can significantly improve inpainting as it does not depend on context sequencing.

We perform the mask diffusion and reverse process in the discrete latent space. After encoding the masked image $x_0 \sim q(\textbf{x}_0)$ to discrete representations $z_{0} \sim q(\textbf{z}_{0})$, we diffuse $\textbf{z}_{0}$ with $t$ steps to $\tilde{\textbf{z}}_{t} \sim q(\textbf{z}_{t}|\textbf{z}_{0})$. Thus the last step with mask $\tilde{\textbf{z}}_{T}^m$ can be demonstrated as $\tilde{\textbf{z}}_{T}^m = (1-m) \times \tilde{\textbf{z}}_{T} + m \times \mathbb{C}$, where $\mathbb{C}\sim \mathrm{Cat}(K,1/K)$ is the sample from a uniform categorical distribution and $m \in \{0,1\}^K $ is the mask, $m=0$ means the context there is masked and $m=1$ means that given the information there. In the reverse process, $\textbf{z}_{T-1}$ can be sampled from $p_{\theta}(\mathbf{z}_{T-1}|\tilde{\textbf{z}}_{T}^m)$ at $t=T$, otherwise, $\textbf{z}_{t-1} \sim p_{\theta}(\mathbf{z}_{t-1}|\textbf{z}_{t}^m)$, and the masked $\textbf{z}_{t-1}^m = (1-m) \times \textbf{z}_{t-1} + m \times \tilde{\textbf{z}}_{t-1}$.

\begin{figure*}[t!]
    \centering{
    \includegraphics[scale=0.145]{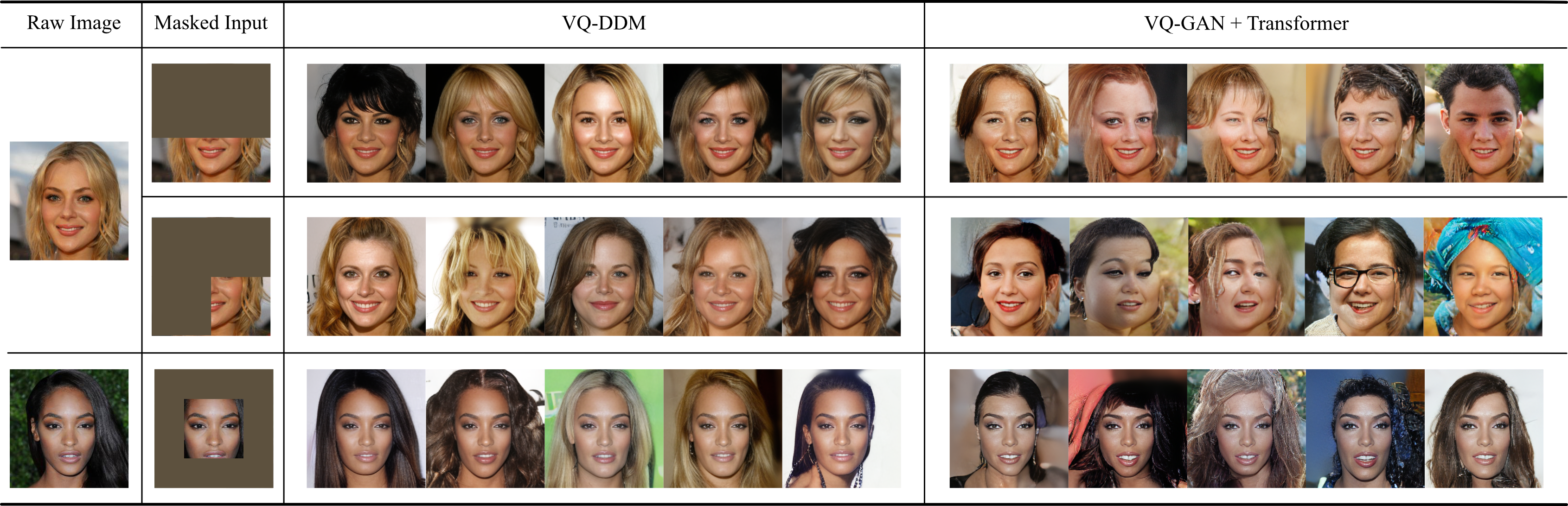}
    }
    \caption{Completions with the arbitrary masks.}
    \label{global_if}
\end{figure*}

 We compare our approach and another that exploits a transformer with a sliding attention window as an autoregressive generative model~\cite{esser2021taming}. The completions are shown in Fig.~\ref{global_if}, in the first row, the upper 62.5\% (160 out of 256 in latent space) of the input image is masked and the lower 37.5\% (96 out of 256) is retained, and in the second row, only a quarter of the image information in the lower right corner is retained as input. We also tried masking in an arbitrary position. In the third row, we masked the perimeter, leaving only a quarter part in the middle. Since the reverse diffusion process captures the global relationships, the image completions of our model performs much better. Our method can make a consistent completions based on arbitrary contexts, whereas the inpainting parts from transformer lack consistency. It is also worth noting that our model requires no additional training in solving the task of image inpainting.

\begin{table}[t!]
    \centering
    
\resizebox{0.46\textwidth}{!}{
    \begin{threeparttable}

\begin{tabular}{llll}
\toprule
Method             & FID  $\downarrow$                  & Params    & FLOPs   \\ \midrule
\multicolumn{2}{l}{\textbf{\textit{Likelihood-based}}}           \\ \midrule
GLOW~\cite{kingma2018glow}                  & 60.9      & 220 M     & 540 G         \\
NVAE~\cite{vahdat2020nvae}                  & 40.3      & 1.26 G    & 185 G       \\
\textbf{\textit{ours}} ($K=1024$ w/o ReFiT)    & 22.6      & 117 M    & 1.06 G \\
VAEBM~\cite{xiao2020vaebm}                  & 20.4      & 127 M    & 8.22 G    \\
\textbf{\textit{ours}} ($K=512$ w/ ReFiT)            & 18.8      & 117 M    & \textbf{1.04 G }     \\
DC-VAE~\cite{parmar2021dual}                & 15.8      & -    & -     \\
\textbf{\textit{ours}} ($K=1024$ w/ ReFiT)           & 13.2      & 117 M    & 1.06 G        \\
DDIM(T=100)~\cite{song2020denoising} &10.9 &114 M &124 G \\
VQ-GAN + Transformer~\cite{esser2021taming}               & 10.2      & 802 M    & 102 G\tnote{a}        \\ \midrule
\multicolumn{2}{l}{\textbf{\textit{Likelihood-free}}} \\\midrule
PG-GAN~\cite{karras2017progressive}             & 8.0   & 46.1 M    & 14.1 G    \\      \bottomrule
\end{tabular}

        \begin{tablenotes}
            \item[a] VQ-GAN is an autoregressive model, and the number in the table is the computation needed to generate the full size latent feature map. The FLOPs needed to generate one discrete index out of 256 is 0.399 G. 
        \end{tablenotes}

    \end{threeparttable}}
    \caption{FID on CelebA HQ $256\times256$ dataset. All the FLOPs in the table only consider the generation stage or inference phase for one $256\times256$ images. }
    \label{celeba}
\end{table}

\section{Related Work}
\subsection{Vector Quantised Variational Autoencoders}
VQ-VAE~\cite{van2017neural} leads a trend of discrete representation of images. The common practice is to model the discrete representations using an autoregressive model, e.g. PixelCNN~\cite{van2016pixel,chen2018pixelsnail}, transformers~\cite{esser2021taming,ramesh2021zero,ramesh2021zero}, etc. 
Some works had attempted to fit the prior distribution of discrete latent variables using a light non-autoregressive approach, like EM approach~\cite{roy2018theory} and Markov chain with self-organizing map~\cite{fortuin2018som}, but yet they are struggling to fit a large scale of data. 
Ho \etal \cite{ho2020denoising} have also shown that the diffusion models can be regarded as an autoregressive model along the time dimension, but in reality, it is non-autoregressive along the pixel dimension.

A concurrent work~\cite{esser2021imagebart} follow a similar pipeline which uses a diffusion model on discrete latent variables, but the work uses parallel modeling of multiple short Markov chains to achieve denoising.

\subsection{Diffusion Models}
Sohl-Dickstein \etal \cite{sohl2015deep} presented a simple discrete diffusion model, which diffused the target distribution into the independent binomial distribution. Recently, Hoogeboom \etal \cite{hoogeboom2021argmax} have extended the discrete model from binomial to multinomial. Further, Austin \etal \cite{austin2021structured} proposed a generalized discrete diffusion structure, which provides several choices for the diffusion transition process.

In the continuous state space, there are some recent diffusion models that surpassed the state-of-the-art in the image generation area. With the guidance from the classifiers, Dhariwal \etal \cite{dhariwal2021diffusion} enabled diffusion models called ADM to generate images beyond BigGAN, which was previously one of the most powerful generative models. In CDM~\cite{ho2021cascaded}, the authors performed the cascade pipeline on the diffusion model to generate the image with ultra-high fidelity and reach state-of-the-art on conditional ImageNet generation.
In addition, there have been several recent works that have attempted to use diffusion models to modelling the latent variables of VAE~\cite{kingma2021variational,wehenkel2021diffusion}, while revealed the connection among several diffusion models mentioned above.
\section{Conclusion}
In this paper, we introduce VQ-DDM, a high-fidelity image generation model with a two-stage pipeline. In the first stage, we train a discrete VAE with a well-utilized content-rich codebook. With the help of such an efficient codebook, it is possible to generate high-quality images by a discrete diffusion model with relatively tiny parameters in the second stage. Simultaneously, benefiting from the discrete diffusion model, the sampling process captures the global information and the image inpainting is no longer affected by the location of the given context and mask. Meanwhile, in comparison with other diffusion models, our approach further reduces the gap in generation speed with respect to GAN.
We believe that VQ-DDM can also be utilized for audio, video and multimodal generation.
\subsection*{Limitations}
For a complete diffusion, we need a large number of steps, which will result in a very fluctuating training process and limit the image generation quality. Hence, our model may suffer from underperformance when exposed to the large scale and complex datasets.

{\small
\bibliographystyle{ieee_fullname}
\bibliography{egbib}
}

\clearpage
\appendix
\section*{Appendix}
\subsection*{Datasets}

\textit{CelebA-HQ} is a high-quality version of the CelebA dataset, consisting of 30000 images generated by PG-GAN. We followed ~\cite{karras2017progressive} instructions to obtain the dataset. 

\textit{LSUN}~\cite{yu2015lsun} includes ten scenes and twenty object categories, totally about one million images with label. We mainly use the \textit{Church}, which contains about 126,000 images. The image pre-processing method follows StyleGAN~\cite{karras2019style}.

\subsection*{Discrete VAEs}
Our architecture for discrete image representation follows that in~\cite{esser2021taming}. For completeness, a brief description is as follows:

\begin{table}[ht]
\centering
\begin{threeparttable}
\begin{tabular}{c|c}
\toprule
Encoder                           & Decoder                           \\ \midrule
Conv2D                            & Conv2D                            \\
4$\times$\{ResDown\} & Middle Block                      \\
Middle Block                      & 4$\times$\{ResDown\} \\
GN, Swish, Conv2D                 & GN, Swish, Conv2D   \\ \bottomrule      
\end{tabular}
\begin{tablenotes}
    \item[1] ResDown is the combination of a Residual Block and Downsample Block, if the feature map size matches the preset value, there will be an addition non-local self-attention block. 
    \item[2] Middle Block is the cascade of one Residual Block, one Self-attention Block and one more Residual Block.
    \item[3] GN means the group normalization~\cite{wu2018group}
\end{tablenotes}
\end{threeparttable}
\caption{Brief Architecture of the VQ-GAN encoder and decoder}
\end{table}

For \textit{CelebA-HQ} and \textit{ImageNet}, we obtain the pre-trained checkpoints from the official release, for \textit{LSUN-Church}, we trained a model from scratch under the same configurations for ImageNet in~\cite{esser2021taming}. Specifically, the embedding dimension is 256 and the number of embedded tokens is 1024. The channel numbers of the encoder-decoders is 128, the self-attention block is introduced when the feature map size meets $16\times16$. We set the learning rate is 4.5e-6 for each instance and the learning rate is fixed.

\subsection*{Discrete Diffusion Models}
The network structures and hyperparameter settings of discrete diffusion models follow~\cite{ho2020denoising}. In detail, the model architecture is based on the backbone of PixelCNN++~~\cite{salimans2017pixelcnn++}, which is a U-Net~\cite{ronneberger2015u} with group normalization. Instead of only adding a self-attention block at $16\times16$ feature map resolution level, we increase two more self-attention blocks on $8\times8$ and $4\times4$ separately. We have 117M parameters for the diffusion models.

For the logits of $\tilde{p_{\theta}}(\tilde{z_0}| z_t) = \mathrm{Cat}(\tilde{z_0}|p_{\theta})$, we predict a noise using the neural network and add it to $z_t$ instead of predicting the $\tilde{z_0}$ directly. As shown in Eq.~\ref{pnois}, the desired logits is obtained by superimposing the predicted noise $\mathrm{nn_{\theta}}(z_t)$ on a calculated $z_t$

The noise schedule $\alpha_t$ is the same as~\cite{nichol2021improved}. The difference is that their parameter $\sqrt{\hat{\alpha_t}}$ is assigned to the mean of the Gaussian distribution, while our factor $\hat{\alpha_t}$ is the parameter of the categorical distribution. The definition is given in Eq.~\ref{noises} and $s=0.008$. We also sample $t$ with $q(t)\propto\sqrt{\mathbb{E}[L^2_t]}$ instead of uniform sampling~\cite{nichol2021improved}.

The batch size is $180$ per GPU and the learning rate is $0.0001$ with Adam optimizer with standard settings. The learning rate scheduler is the cosine annealing scheduler with 1 million steps. We have not employed any dropout in the model.

\subsection*{Additional Results}

\begin{figure*}
    \centering
    \includegraphics[scale=0.25]{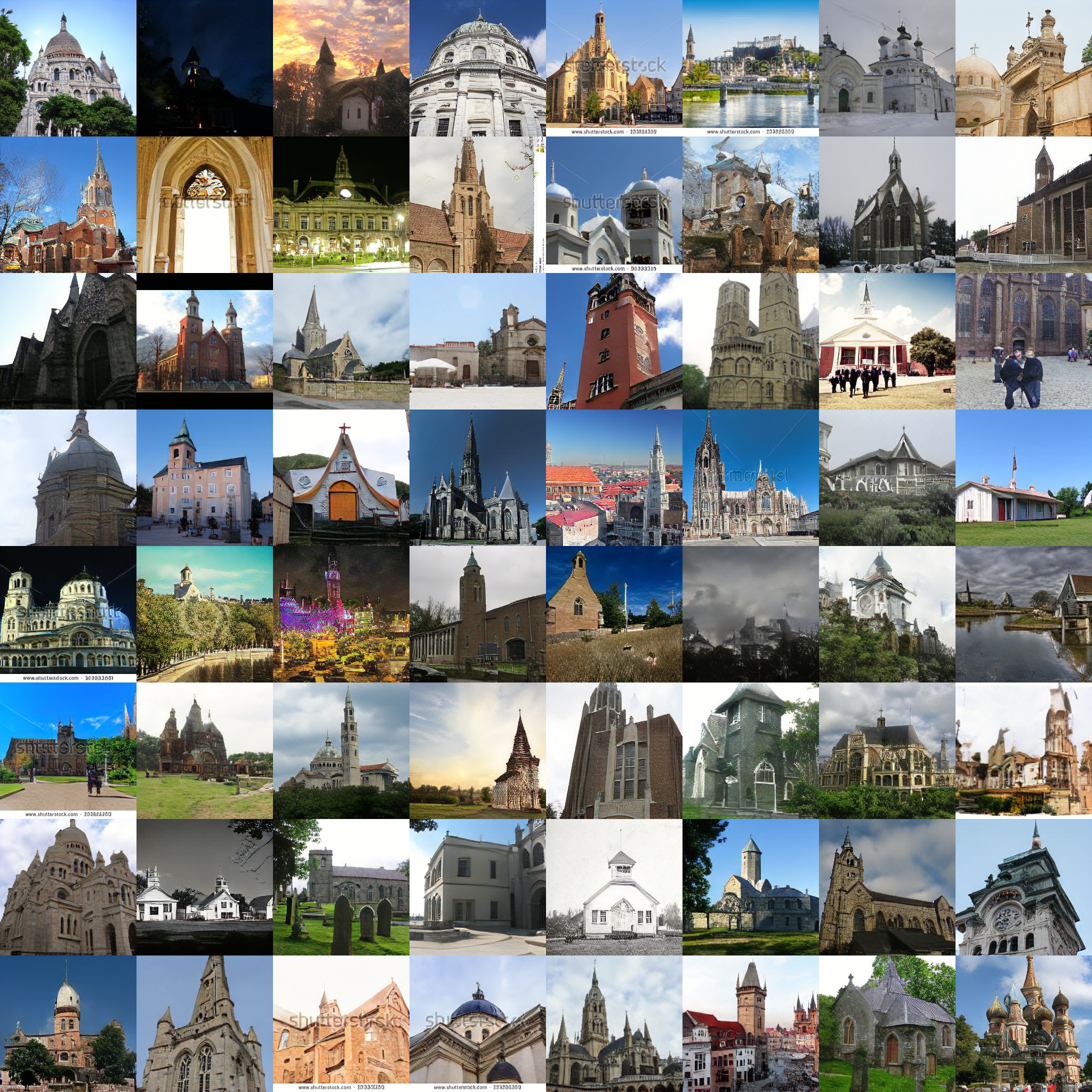}
    \caption{Additional samples on LSUN-Church.}
    \label{addfigs_lc}
\end{figure*}

\begin{figure*}
    \centering
    \includegraphics[scale=0.25]{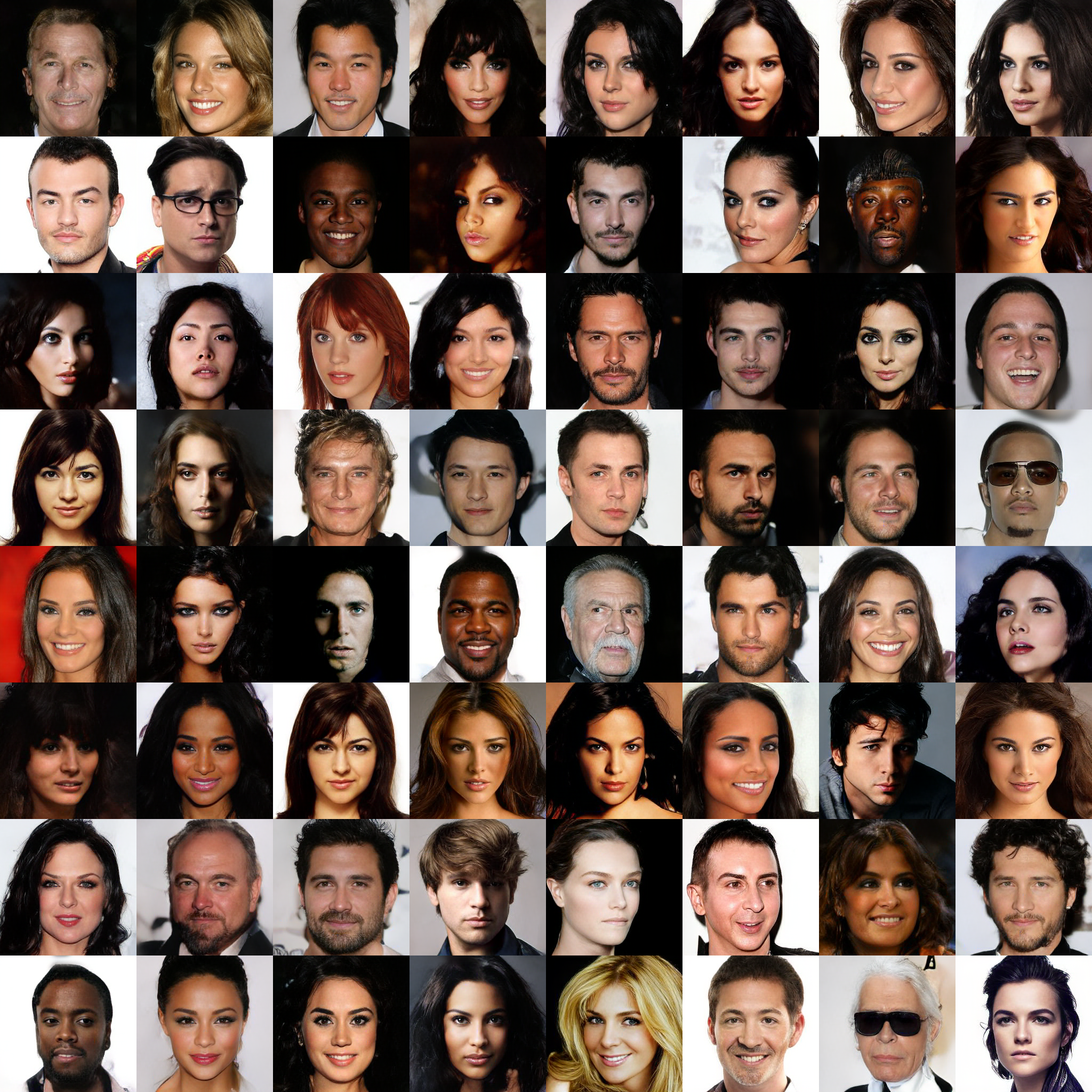}
    \caption{Additional samples on CelebA-HQ.}
    \label{addfigs_ce}
\end{figure*}

\begin{figure*}
    \centering
    \includegraphics[scale=0.25]{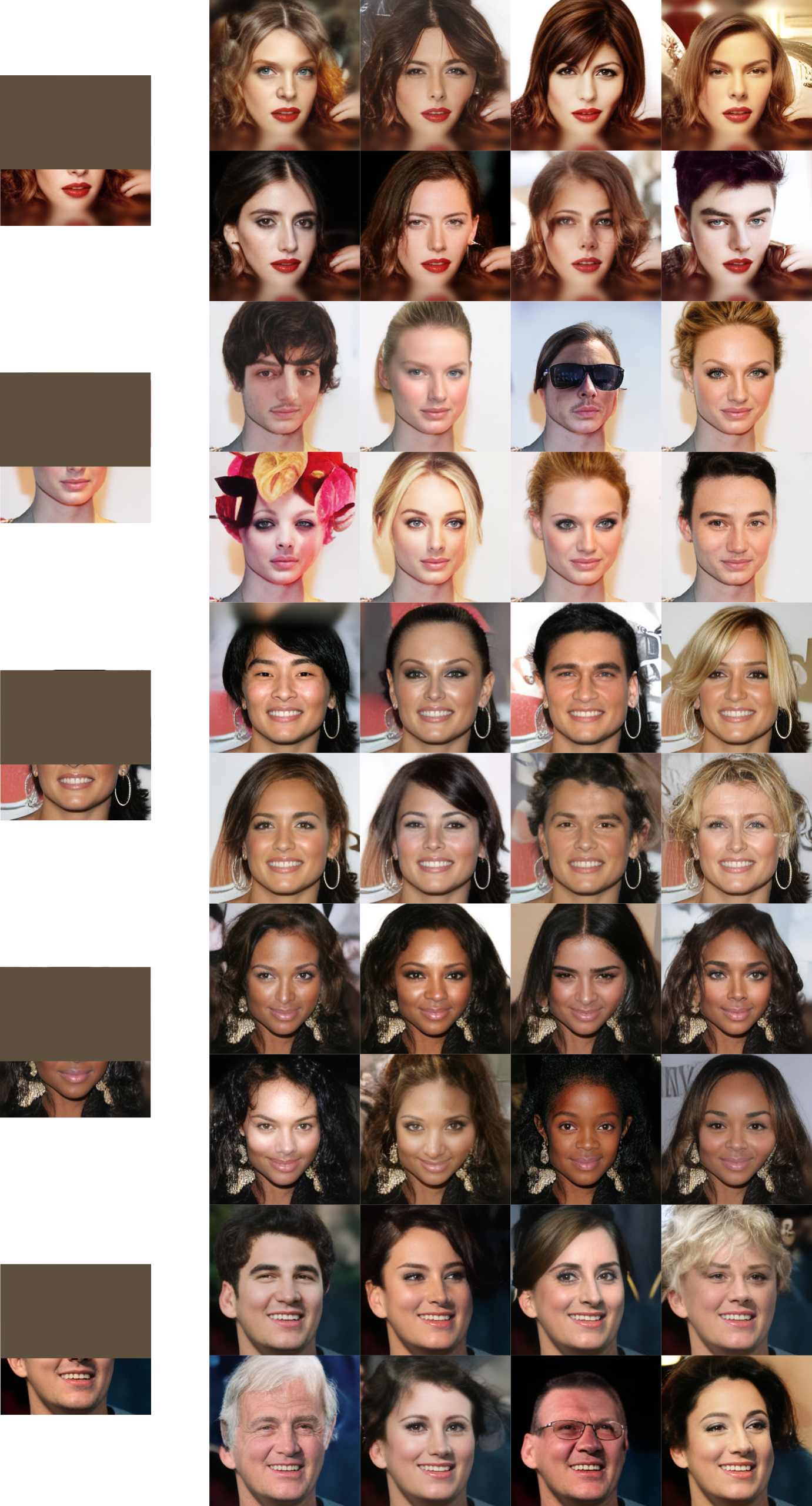}
    \caption{Additional samples on image inpainting for CelebA-HQ.}
    \label{addfigs_gb}
\end{figure*}

In Figures~\ref{addfigs_lc} \& \ref{addfigs_ce}, we show additional generation results based on CelebA and on LSUN-Church. We also provide additional results for image inpainting in Fig.~\ref{addfigs_gb}.

\subsection*{Risk of overfitting}

\begin{figure*}
    \centering
    \includegraphics[scale=0.17]{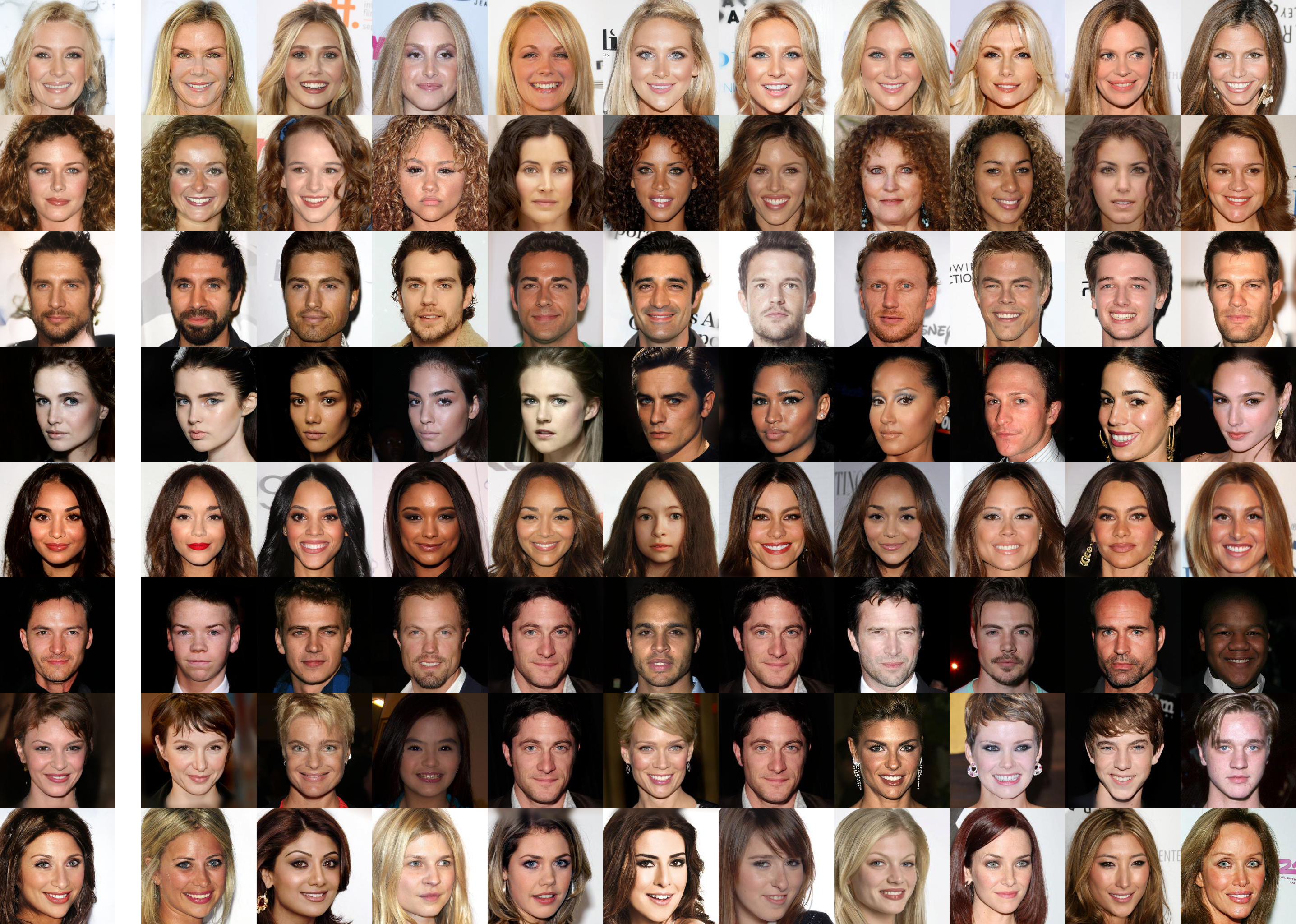}

    \caption{Nearest Neighbours for CelebA-HQ $256\times256$ model. The left column are the images generated by our model, and the remaining images are the nearest neighbors(with minimum LPIPS distance) from the training set.}
    \label{nns}
\end{figure*}

As described in \cite{esser2021taming}, FID scores cannot detect an overfitting, while early-stopping based on validation NLL can prevent overfitting. In Fig~\ref{nns}, we show top-$10$ nearest neighbors based on \textit{LPIPS} distance~\cite{zhang2018perceptual} for the training image. We can find that the nearest neighboring generated image is not the reproduced original image and we can infer that there is no overfitting in such model.

\subsection*{Societal Impact}
Our work is an extension of the diffusion model, which also belongs to the family of generative models. It can be used to generate fake images or videos to disseminate disinformation, however, as our adopted datasets are collected from the Internet, which will contain the biases, the generated images from our model are also difficult to escape from the bias caused by training data.




\end{document}